\documentclass[11pt]{article}
\usepackage{geometry}
\usepackage{coling2020}
\usepackage{times}
\usepackage{url}
\usepackage{latexsym}
\usepackage{amsmath}
\usepackage{amssymb}
\usepackage{microtype}
\usepackage{graphicx}
\usepackage{hyperref}
\usepackage{multirow}
\usepackage{caption, subcaption}
\usepackage{booktabs}

\colingfinalcopy % Uncomment this line for all SemEval submissions

\title{ISCAS at SemEval-2020 Task 5: Pre-trained Transformers for Counterfactual Statement Modeling}

\author{Yaojie Lu${}^{1,3}$, Annan Li${}^{1,3}$, Hongyu Lin${}^{1}$, Xianpei Han${}^{1,2}$, Le Sun${}^{1,2}$ \\
${}^{1}$Chinese Information Processing Laboratory ~ ${}^{2}$State Key Laboratory of Computer Science \\
Institute of Software, Chinese Academy of Sciences, Beijing, China\\
${}^{3}$University of Chinese Academy of Sciences, Beijing, China \\
 {\tt \{yaojie2017,liannan2019,hongyu2016,xianpei,sunle\}@iscas.ac.cn}
}

\date{}

\begin{document}
\maketitle
\begin{abstract}
  ISCAS participated in two subtasks of SemEval 2020 Task 5: detecting counterfactual statements and detecting antecedent and consequence.
  This paper describes our system which is based on pre-trained transformers.
  For the first subtask, we train several transformer-based classifiers for detecting counterfactual statements.
  For the second subtask, we formulate antecedent and consequence extraction as a query-based question answering problem.
  The two subsystems both achieved third place in the evaluation.
  Our system is openly released at \href{https://github.com/casnlu/ISCAS-SemEval2020Task5}{https://github.com/casnlu/ISCAS-SemEval2020Task5}.
\end{abstract}

\section{Introduction}
\label{intro}

Counterfactual statements describe events that did not actually happen or cannot occur, as well as the possible consequence if the events have had happened.
Counterfactual detecting aims to identify counterfactual statements in language and understand antecedents and consequents in these statements.
For instance, the following sentence is a counterfactual statement, and the underlined term is the \underline{antecedent}, while the italic term is the \textit{consequence}:
\begin{quote}
  \underline{Her post-traumatic stress could have been avoided} \textit{if a combination of paroxetine and exposure therapy had been prescribed two months earlier}.
\end{quote}
Once understanding the statement, we can accumulate the causal knowledge for ``post-traumatic stress'', i.e., ``a combination of paroxetine and exposure may help cure post-traumatic stress''.
To model counterfactual semantics and reason in natural language, SemEval 2020 Subtask 5 provides an English benchmark for two basic problems: detecting counterfactual statements and detecting antecedent and consequence \cite{yang-2020-semeval-task5}.

We build our evaluation systems that are built on pre-trained transformer-based neural network models, which have shown significant improvements over conventional methods in many NLP fields \cite{devlin-etal-2019-bert,liu2020roberta,Lan:ICLR2020:ALBERT}.
Specifically, in subtask 1, several transformer-based classifiers are designed to detect counterfactual statements.
Besides, because counterfactual antecedent expressions are usually expressed using some obvious conditional assumption connectives, such as \textit{if} and \textit{wish}.
We also equip transformers with additional convolutional neural network to capture the above strong local context information.
For subtask 2, we formulate antecedent and consequence extraction as a query-based question answering problem.
Specifically, to effectively model context information in counterfactual statements, we design two different kinds of input queries for antecedents/consequences and regard counterfactual statements as given paragraphs.

The rest of this paper is organized as follows.
Section \ref{background} introduces the background of pre-trained transformers.
Section \ref{system} describes the overview of our system for two subtasks.
In Section \ref{setup}-\ref{results}, we describe the detailed experiment setup and the overall system performance on the two subtasks.
Finally, we conclude this paper in Section \ref{conclusion}.

\blfootnote{
    \hspace{-0.65cm}  % space normally used by the marker
    This work is licensed under a Creative Commons 
    Attribution 4.0 International Licence.
    Licence details:
    \url{http://creativecommons.org/licenses/by/4.0/}.
}

\section{Background}
\label{background}

Different from the pre-trained word embedding in NLP \cite{pennington-etal-2014-glove}, pre-trained contextualized models aim to learn encoders to represent words in context for downstream tasks.
BERT\cite{devlin-etal-2019-bert} is a representative large-scale pre-trained transformer, which is trained using mask language modeling (MLM) and next sentence prediction (NSP) task.

Whole Word Masking model\footnote{https://github.com/google-research/bert} (BERT-WWM) is a simple but effective variant of BERT.
In this case, the pre-training stage always mask all of the tokens corresponding to a word instead of a single WordPiece token (sub-token).

RoBERTa \cite{liu2020roberta} further improves on BERT's pre-training procedure and achieves substantial improvements. The improvements include training the model longer, with bigger batches over more data; removing the next sentence prediction objective; training on longer sequences; and dynamically changing the masking pattern applied to the training data.

ALBERT \cite{Lan:ICLR2020:ALBERT} incorporates factorized embedding parameterization and cross-layer parameter sharing to reduce the number of parameters of BERT.
These two methods can significantly reduce the number of parameters of BERT, thus improving the parameter efficiency and facilitating the learning of larger models.
Besides, ALBERT also uses sentence ordering prediction (SOP) self-supervised learning task to replace BERT's NSP task, for it’s helpful for the model to better learn sentence coherence.

\section{System overview}
\label{system}

Given a candidate text $x=\{w_{1}, w_{2}, ..., w_{n}\}$, our system needs to: 1) determine whether the candidate contains a counterfactual statement; 2) extract the antecedent and consequence from the counterfactual statement.
For the example in Section \ref{intro}, we first detect it is a counterfactual statement, then extract ``\textit{Her post-traumatic stress could have been avoided}'' as its antecedent and ``\textit{if a combination of paroxetine and exposuretherapy had been prescribed two months earlier}'' as its consequent.
In the following, we describe our two sub-systems in detail.

\subsection{Detecting Counterfactual Statements as Text Classification}
\label{subtask1}

\begin{figure}[!tpb]
  \centering
  \includegraphics[width=0.96\textwidth]{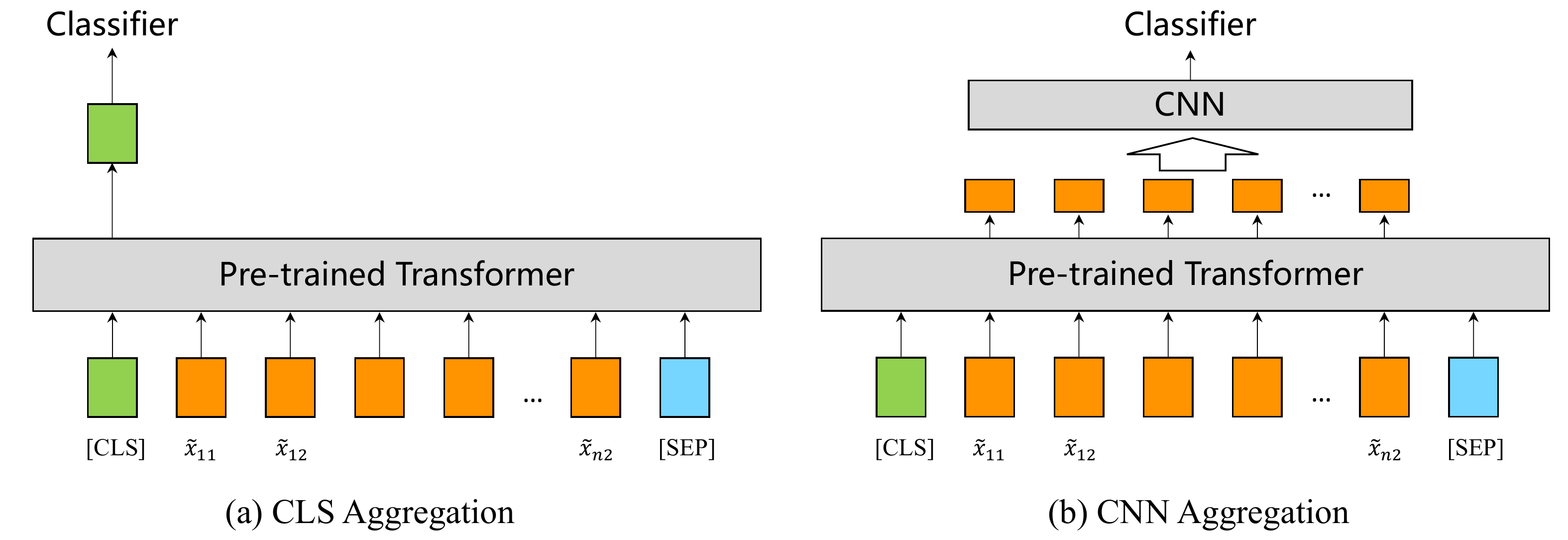}
  \caption{The Transformer for Detecting Counterfactual Statements.}
  \label{fig:framework_subtask1}
  \end{figure}

To detect counterfactual statements, we build classifiers based on contextualized representation to detect counterfactual statements.
We first represent each word in the text using its contextualized representation, then obtain the overall text representation using two different aggregation methods, and finally determining whether the text contains counterfactual statements using a classifier.
The overall framework is shown in Figure \ref{fig:framework_subtask1}.

\paragraph{Contextualized Word Representation Layer.}

To capture the counterfactual semantics in natural language, we learn a contextualized representation for each token.
In order to alleviate the out-of-vocabulary problem in text representation, we first convert the raw input text into word-pieces (sub-tokens) $\{\tilde{x}_{11}, ..., \tilde{x}_{n1}, \tilde{x}_{n2}\}$ in the pre-defined vocabulary.
Then, the two special symbols [CLS] and [SEP] will be added to the head and tail of the sentence.
Finally, we feed tokenized text $\tilde{\textbf{x}}=\{\text{[CLS]}, \tilde{x}_{11}, ..., \tilde{x}_{n1}, \tilde{x}_{n2}, \text{[SEP]}\}$ into $L$-layers pre-trained transformers to obtain the contextualized representation for each sub-tokens.

Following \cite{tenney-etal:2019:iclr2019}, we pool token $i$'s representation $\mathbf{h}_{i} \in \mathbb{R}^{n \times d}$ across all BERT layers using scalar mixing \cite{peters-etal:2018:naacl2018}:
$\mathbf{h}_{i} = \gamma \sum_{j=1}^{L} \alpha_{j} \mathbf{x}_{i}^{(j)}$
where $\textbf{x}_{i}^{(j)}\in \mathbb{R}^{d}$ is the embedding of token $i$ from BERT layer $j$, $\alpha_{j}$ is softmax-normalized weights, and $\gamma$ is a scalar parameter.
We denote the final representation of the special symbol [CLS] as $C \in \mathbb{R}^{d}$.
Specifically, we obtain the token-level representation using the representation of the first sub-token in each token.

\paragraph{Contextualized Information Aggregation.}

After obtaining the representation of each word, we produce aggregated feature vector $\mathbf{r}$ to capture the counterfactual information of the entire statement.
We investigate two different aggregation strategies in this section: [CLS] aggregation and convolutional neural network (CNN) aggregation.

In [CLS] aggregation, we directly use the representation $C$ of the special symbol [CLS] as the aggregate feature $\mathbf{r}$ \cite{devlin-etal-2019-bert}.

In counterfactual statements, connectives are often used to express the relation between antecedent and consequence, i.e., ``if'', ``even if'', and ``would''.
To capture these local patterns in counterfactual statements, we employ a CNN \cite{kim-2014-convolutional} to aggregate sentence information.
Given the token sequence $\{\mathbf{h}_{1}, ..., \mathbf{h}_{n}\}$, the convolutional filter scans the token sequence and extract the local feature $\mathbf{l}_{i}$: $\mathbf{l}_{i} = \tanh \mathbf{w} \cdot \mathbf{h}_{i:i+h-1} + \mathbf{b}$.
Finally, a max-pooling layer is used to produce the feature $\mathbf{r}$ for further counterfactual statement detection: $\mathbf{r}=\max_{0 \leq i \leq n} \mathbf{l}_{i}$.

\paragraph{Counterfactual Statement Classifier.}

After aggregation, the feature vector $\mathbf{r}$ will be fed to the counterfactual classifier, which computes a probability  of whether it is a counterfactual statement:
\begin{equation}
  P(y=1|x) = \sigma(\mathbf{w}_{c}\cdot \mathbf{r} + b_{c})
\end{equation}
where $\mathbf{w}_{c}$ is the weight vector, $b_{c}$ is the bias term, and $\sigma$ is simgoid function.

Given the training set $D=\{(x_{i}, y_{i})\}$, we train all parameters using a binary cross-entropy loss function:
\begin{equation}
  \mathcal{L} = \sum_{i \in \mathcal{D}} y_{i} \log P(y=1|x_{i})  + (1 - y_{i}) \log (1 - P(y=1|x_{i}))
\end{equation}

\subsection{Extracting Antecedent and Consequence as Question Answering}
\label{subtask2}

\begin{figure}[!tpb]
  \centering
  \includegraphics[width=0.50\textwidth]{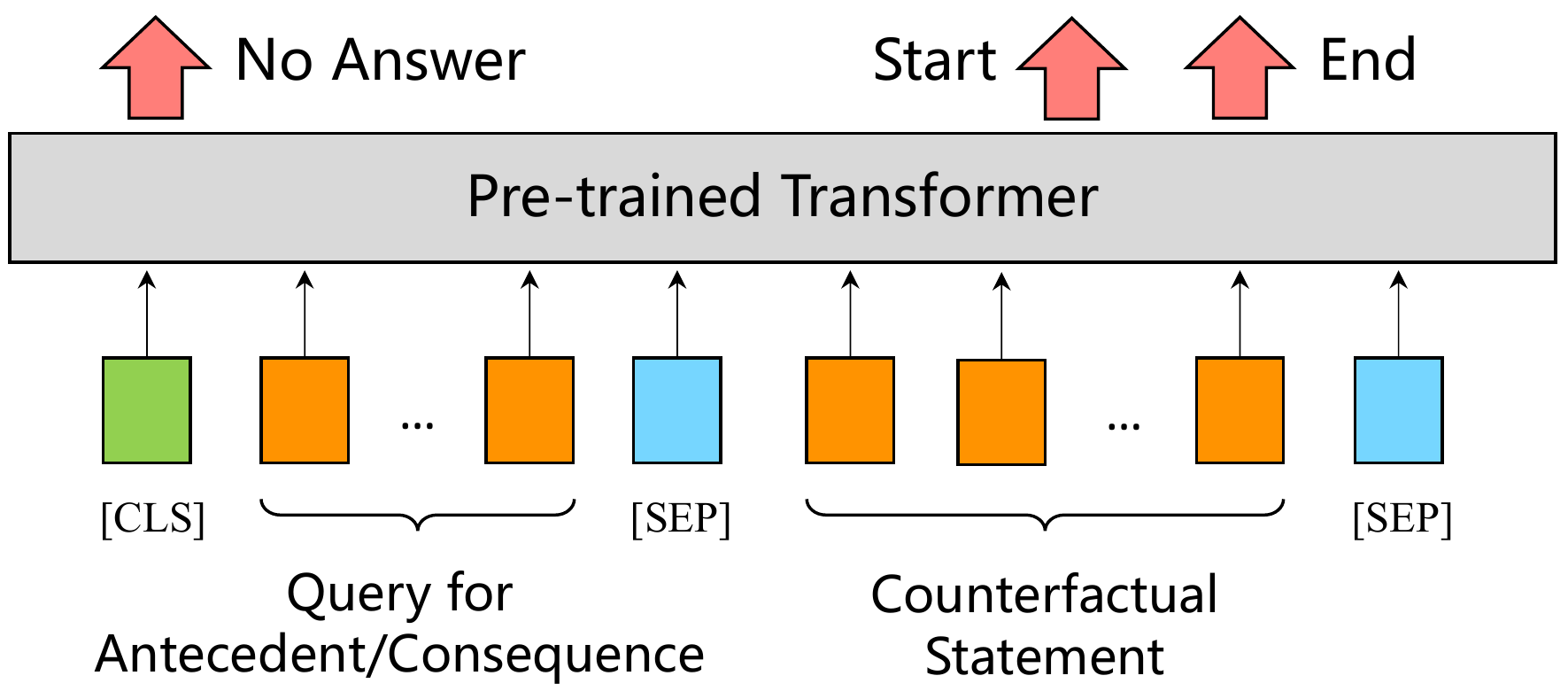}
  \caption{The Transformer for Detecting Antecedent and Consequence.}
  \label{fig:framework_subtask2}
  \end{figure}

We now describe how to extract antecedent and consequence via a question answering-style procedure.
Given a counterfactual statement $s$, we first construct an antecedent query $q_{a}$ and a consequence query $q_{c}$ separately, and then extract the corresponding antecedent $a_{a}$ and consequence $a_{c}$ in the text by answering these two questions.
The overall framework is illustrated in Figure \ref{fig:framework_subtask2}.

\paragraph{Query Construction.}
We design two kinds of queries for extraction: name query and definition query.
For name query, we directly use ``antecedent'' and ``consequence'' as the query for extraction.
To enrich the semantic information of questions, we also propose definition query, which employs the dictionary definition\footnote{https://www.merriam-webster.com} of each label as definition queries.
For ``antecedent'', the definition query is ``a preceding event, condition, or cause''.
For ``consequent'', the definition query is ``a result or effect''.

\paragraph{Question and Context Encoding.}

We represent the input question $q_{*}$ for extraction and the counterfactual statement $s$ as a single packed sequence: \{[CLS], $q_{*}$, [SEP], $s$, [SEP]\}.
First, $q_{*}$ and $s$ are tokenized as sub-token sequences after WordPiece tokenization as shown in Figure \ref{fig:framework_subtask2}.
After tokenization, we feed the single packed sequence to the pre-trained transformers, and obtain the final hidden vector for the $i^{\text{th}}$ sub-token in the query as $\mathbf{h}_{q}^{i} \in \mathbb{R}^{d}$, the $j^{\text{th}}$ sub-token in the statement as $\mathbf{h}_{s}^{j} \in \mathbb{R}^{d}$, and $C  \in \mathbb{R}^{d}$ for the special token [CLS].

\paragraph{Answer Prediction.}

To extract continuous text fragments, we employ a pointer network to predict the start position and end position of the answer text.
A pointer network contains a start vector $\mathbf{w}_{\text{start}}$ and an end vector $\mathbf{w}_{\text{end}}$, which are used to produce the scores of word $i$ being the start/end of the answer.
The score of word $i$ being the start of the answer is computed as a dot product between the hidden state of each token in the statement: $\mathbf{w}_{\text{start}} \cdot \mathbf{h}_{s}^{j}$, the score of the end is calculated  in the same way.
We define the score of a candidate span from position $j$ to position $k$ as $S_{j,k} = \mathbf{w}_{\text{start}} \cdot \mathbf{h}_{s}^{j} + \mathbf{w}_{\text{end}} \cdot \mathbf{h}_{s}^{k}$, where $k \geq j$.

Since some statements do not contain consequences\footnote{These statements cover 14.64\% in the training set.}, we regard the questions corresponding to these consequence statements as unanswerable questions.
For these questions, we treat [CLS] token as both the start and the end of the answer span.
In this way, the score of a statement without consequence is $S_{\text{null}} = \mathbf{w}_{\text{start}} \cdot C + \mathbf{w}_{\text{end}} \cdot C$.

For model training, we update the full model by maximizing the likelihood of the start token $j^{*}$ and the end token $k^{*}$ (including [CLS]):
\begin{equation}
  \begin{gathered}
    \mathcal{L} = - \sum_{i \in \mathcal{D}} \log P(y_{start}=j^{*}|x_{i}) + \log P(y_{end}=k^{*}|x_{i}) \\
    P(y_{start}=j^{*}|x_{i}) = \frac{\exp (\mathbf{w}_{\text{start}} \cdot \mathbf{h}_{s}^{j^{*}})}{\exp (\mathbf{w}_{\text{start}} \cdot C) + \sum_{j=1}^{n} \exp (\mathbf{w}_{\text{start}} \cdot \mathbf{h}_{s}^{j}) } \\
    P(y_{end}=k^{*}|x_{i}) = \frac{\exp (\mathbf{w}_{\text{end}} \cdot \mathbf{h}_{s}^{k^{*}})}{\exp (\mathbf{w}_{\text{end}} \cdot C) + \sum_{k=1}^{n} \exp (\mathbf{w}_{\text{end}} \cdot \mathbf{h}_{s}^{k}) }
  \end{gathered}
\end{equation}
where the parameters of the pointer network are training from scratch.

\section{Experimental setup}
\label{setup}

\subsection{Data Splits}
\label{sec:data_splits}

\paragraph{Subtask 1.}
This subtask contains 13,000 instances for model training and 7000 unseen instances for online evaluation.
We sampled 1,500 instances from the whole dataset as our development set.
Then, we split the remaining 11,500 instances in 5-fold; each fold has 2,300 instances.
We trained five models on five groups of datasets for ensemble voting.
Each group takes four folds as a training dataset and the remaining fold for early stopping.

\paragraph{Subtask 2.}
This subtask contains a total of 3,551 instances for model training and 1950 unseen instances for online evaluation.
We sampled 3200 instances as training sets and take the remaining 351 instances as development sets.

\subsection{Implementation and Hyperparameters}
\paragraph{Subtask 1.} 
For each model, we selected the best fine-tuning learning rate (among `5e-6' `1e-5' `3e-5') on the development set.
Because of the GPU memory limitation, we truncated the maximum total input sequence length after WordPiece tokenization to 128.
We employ three different pre-trained transformers in our submission for SemEval 2020 Task 5 official evaluation: BERT \cite{devlin-etal-2019-bert}, ALBERT \cite{Lan:ICLR2020:ALBERT}, and RoBERTa \cite{liu2020roberta}.
We used a batch size of 8 for ALBERT-xxlarge and 24 for other models.
For CNN aggregation, we only used one single  CNN layer whose window size is 3 and hidden size is 300.

\paragraph{Subtask 2.}
We fine-tuned all models on the training data for 5 epochs using the learning rate of $1 \times 10^{-5}$ for the BERT parameters and the task parameters, while we evaluate and save models at every 250 steps with the batch size of each step is 16.
We trained the large model on two 24G GPUs in parallel, and selected the best model on the development set for online evaluation.

\section{Results}
\label{results}

We report the performance of subtask 1 and subtask 2, scored by the evaluation server\footnote{https://competitions.codalab.org/competitions/21691}.
In subtask 1, different models are evaluated using Precision ($P$), Recall ($R$), and F1-score ($F_{1}$) for binary classification.
While there are four metrics for subtask 2: Exact Match ($EM$), Precision, Recall, and F1-score.
Exact match measures the percentage of predictions that match the annotated antecedents and consequences exactly.
Note that, F1-score in subtask 2 is token level metric and will be calculated based on the offsets of predicted antecedent and consequence.

\begin{table}[htbp]
  \centering
  \resizebox{0.50\textwidth}{!}{
    \begin{tabular}{cccc}
      \toprule
          & $F_{1}$    & $R$     & $P$ \\
      \midrule
    $\text{BERT}_{\text{Large-Cased-WWM}}$ + [CLS] & 87.70 & 87.50 & 87.90 \\
    $\text{BERT}_{\text{Large-Cased-WWM}}$ + CNN & 88.00 & 87.90 & 88.10 \\
    $\text{Roberta}_{\text{Large}}$ + [CLS] & 89.80 & \textbf{90.40} & 89.20 \\
    $\text{Roberta}_{\text{Large}}$ + CNN & 89.70 & 89.60 & 89.80 \\
    $\text{ALBERT}_{\text{XXLarge}}$ + [CLS] & \textbf{90.00} & 87.90 & \textbf{92.20} \\
    $\text{ALBERT}_{\text{XXLarge}}$ + CNN & 89.00 & 87.70 & 90.40 \\
    $\text{ALBERT}_{\text{XXLarge}}$ + [CLS] + CNN & \textbf{90.00} & 88.60 & 91.50 \\
    \bottomrule
    \end{tabular}
  }
  \caption{
      Subtask 1 Test results.
      }
  \label{tab:subtask1-test}
\end{table}

Table \ref{tab:subtask1-test} shows the overall results of our seven runs on subtask 1.
We can see that our system achieved very competitive performance .
The performance of $\text{ALBERT}_{\text{XXLarge}}$ with [CLS] aggregation on precision (92.20) ranked 1$^{st}$ in all teams.
Our best $F_{1}$ (90.00) score ranked 3$^{rd}$ in all teams.

\begin{table}[htbp]
  \centering
  \resizebox{0.60\textwidth}{!}{
    \begin{tabular}{ccccc}
      \toprule
          & $F_{1}$    & $R$     & $P$     & $EM$ \\
    \midrule
    $\text{BERT}_{\text{Base-Cased}}$ + Name & 86.30 & 90.30 & 86.00 & 51.60 \\
    $\text{BERT}_{\text{Base-Uncased}}$ + Name & 86.60 & 90.20 & 86.70 & 51.90 \\
    $\text{BERT}_{\text{Base-Cased}}$ + Definition & 86.30 & 90.30 & 86.00 & 52.40 \\
    $\text{BERT}_{\text{Base-Uncased}}$ + Definition & 86.80 & 90.00 & 87.10 & 52.50 \\
    \hline
    $\text{BERT}_{\text{Large-Uncased-WWM}}$ + Name & 87.30 & 89.80 & \textbf{87.80} & 54.40 \\
    $\text{BERT}_{\text{Large-Uncased-WWM}}$ + Definition & \textbf{87.50} & \textbf{90.80} & 87.50 & \textbf{54.60} \\
    \bottomrule
    \end{tabular}
  }
  \caption{
    Subtask 2 Test results.
    Name indicates using name queries and Def. indicates using definition-enriched queries.
  }
  \label{tab:subtask2-test}
\end{table}

Table \ref{tab:subtask2-test} shows the overall results of our seven runs on subtask 2.
Our QA-based method ranked 1$^{st}$ on $R$ score and 3$^{rd}$ on $F_{1}$, $P$ scores.
Besides, our system achieved 2$^{nd}$ on $EM$ score and surpassed the third team by a large margin (4.90).
From the results in Table \ref{tab:subtask2-test}, we can see that:

1) The definition-based query achieved better performance than the name-based queries. We believe this because the definition-based query provides richer semantic information than the name-based query.

2) Uncased models are better than cased models on both $F_{1}$ and $EM$ scores.
This may be because our model focuses more on capturing the structure information of counterfactual expressions, meanwhile case information is more useful on capturing information about named entities, such as persons and locations.

\section{Conclusion}
\label{conclusion}
In this paper, we propose a transformers-based system for counterfactual modeling.
For counterfactual statements detection, we investigated a variety of advanced pretraining models and two efficient aggregation algorithms.
For antecedent and consequent extraction, we framed it as a span-based question answering task, and then definition-enriched queries are designed to extract the required term from counterfactual statements.
Evaluation results demonstrate the effectiveness of our system.
For future work, we plan to investigate how to inject extra-knowledge into counterfactual modeling systems, such as knowledge-enriched transformers.

\bibliographystyle{coling}
\bibliography{
  semeval2020_subtask5
  }

\begin{thebibliography}{}

\bibitem[\protect\citename{Devlin \bgroup et al.\egroup
  }2019]{devlin-etal-2019-bert}
Jacob Devlin, Ming-Wei Chang, Kenton Lee, and Kristina Toutanova.
\newblock 2019.
\newblock {BERT}: Pre-training of deep bidirectional transformers for language
  understanding.
\newblock In {\em Proceedings of the 2019 Conference of the North {A}merican
  Chapter of the Association for Computational Linguistics: Human Language
  Technologies, Volume 1 (Long and Short Papers)}, pages 4171--4186,
  Minneapolis, Minnesota, June. Association for Computational Linguistics.

\bibitem[\protect\citename{Kim}2014]{kim-2014-convolutional}
Yoon Kim.
\newblock 2014.
\newblock Convolutional neural networks for sentence classification.
\newblock In {\em Proceedings of the 2014 Conference on Empirical Methods in
  Natural Language Processing ({EMNLP})}, pages 1746--1751, Doha, Qatar,
  October. Association for Computational Linguistics.

\bibitem[\protect\citename{Lan \bgroup et al.\egroup
  }2020]{Lan:ICLR2020:ALBERT}
Zhenzhong Lan, Mingda Chen, Sebastian Goodman, Kevin Gimpel, Piyush Sharma, and
  Radu Soricut.
\newblock 2020.
\newblock Albert: A lite bert for self-supervised learning of language
  representations.
\newblock In {\em International Conference on Learning Representations}.

\bibitem[\protect\citename{Liu \bgroup et al.\egroup }2020]{liu2020roberta}
Yinhan Liu, Myle Ott, Naman Goyal, Jingfei Du, Mandar Joshi, Danqi Chen, Omer
  Levy, Mike Lewis, Luke Zettlemoyer, and Veselin Stoyanov.
\newblock 2020.
\newblock Roberta: A robustly optimized bert pretraining approach.

\bibitem[\protect\citename{Pennington \bgroup et al.\egroup
  }2014]{pennington-etal-2014-glove}
Jeffrey Pennington, Richard Socher, and Christopher Manning.
\newblock 2014.
\newblock {G}love: Global vectors for word representation.
\newblock In {\em Proceedings of the 2014 Conference on Empirical Methods in
  Natural Language Processing ({EMNLP})}, pages 1532--1543, Doha, Qatar,
  October. Association for Computational Linguistics.

\bibitem[\protect\citename{Peters \bgroup et al.\egroup
  }2018]{peters-etal:2018:naacl2018}
Matthew Peters, Mark Neumann, Mohit Iyyer, Matt Gardner, Christopher Clark,
  Kenton Lee, and Luke Zettlemoyer.
\newblock 2018.
\newblock Deep contextualized word representations.
\newblock In {\em Proceedings of the 2018 Conference of the North {A}merican
  Chapter of the Association for Computational Linguistics: Human Language
  Technologies, Volume 1 (Long Papers)}, pages 2227--2237, New Orleans,
  Louisiana, June. Association for Computational Linguistics.

\bibitem[\protect\citename{Tenney \bgroup et al.\egroup
  }2019]{tenney-etal:2019:iclr2019}
Ian Tenney, Patrick Xia, Berlin Chen, Alex Wang, Adam Poliak, R.~Thomas McCoy,
  Najoung Kim, Benjamin~Van Durme, Samuel~R. Bowman, Dipanjan Das, and Ellie
  Pavlick.
\newblock 2019.
\newblock What do you learn from context? probing for sentence structure in
  contextualized word representations.
\newblock In {\em 7th International Conference for Learning Representations}.

\bibitem[\protect\citename{Yang \bgroup et al.\egroup
  }2020]{yang-2020-semeval-task5}
Xiaoyu Yang, Stephen Obadinma, Huasha Zhao, Qiong Zhang, Stan Matwin, and
  Xiaodan Zhu.
\newblock 2020.
\newblock {S}em{E}val-2020 task 5: Counterfactual recognition.
\newblock In {\em Proceedings of the 14th International Workshop on Semantic
  Evaluation (SemEval-2020)}, Barcelona, Spain.

\end{thebibliography}

\end{document}